\documentclass[journal]{IEEEtran}
\usepackage[T1]{fontenc}
\usepackage{graphicx}
\usepackage[table]{xcolor}
\usepackage{times}
\usepackage{helvet}
\usepackage{courier}
\usepackage{amsmath}
\usepackage{algorithmic}
\usepackage{csquotes} 
\usepackage{paralist}
\usepackage{amssymb}
\usepackage{indentfirst}
\usepackage{subfigure}
\usepackage{float}
\usepackage{multirow}
\usepackage{cite}
\usepackage{mathrsfs}

\usepackage{mathrsfs} 
\usepackage{amsfonts}
\usepackage{todonotes}
\usepackage{pgfplots} 
\usepackage{epstopdf}
\usepackage{epsfig}
\pgfplotsset{compat=newest}
\definecolor{forestgreen}{RGB}{0,139,69}
\usepackage{multirow}

\definecolor{citecolor}{HTML}{0071bc}
\usepackage{tabularray}
\usepackage{algorithm}
\usepackage{amssymb}
\usepackage{textcomp} 
\usepackage{siunitx}  
\usepackage[colorlinks, linkcolor=red,  anchorcolor=blue, citecolor=citecolor]{hyperref} 

\definecolor{SeaGreen4}{RGB}{0,205,102} 
\definecolor{SlateBlue}{RGB}{106,90,205} 
\definecolor{DarkRed}{RGB}{178,34,34} 

\usepackage{caption}
\usepackage[switch]{lineno}

\usepackage{amsmath}        
\usepackage{booktabs}
\usepackage{hhline}
\usepackage{float}
\usepackage{algorithm}
\usepackage{algorithmic}
\usepackage{makecell}
\usepackage{array}

\makeatletter
\newcommand\fs@ruledred{%
  \def\@fs@cfont{\bfseries}%
  \let\@fs@capt\floatc@ruled
  \def\@fs@pre{{\color{black}\hrule height.8pt depth0pt \kern2pt}}%
  \def\@fs@post{{\color{black}\kern2pt\hrule}}%
  \def\@fs@mid{{\color{black}\kern2pt\hrule\kern2pt}}%
  \let\@fs@iftopcapt\iftrue
}
\makeatother

\floatstyle{ruledred}
\restylefloat{algorithm}

\usepackage{textcomp,booktabs}
\usepackage{amssymb}
\usepackage{pifont}

\usepackage{makecell}
\usepackage{array}
\usepackage{colortbl}
\definecolor{mygray}{gray}{.9}
\definecolor{mypink}{rgb}{.99,.91,.95}
\definecolor{mycyan}{cmyk}{.3,0,0,0}

\begin{document}


\title{FusionCLIP: Multimodal Contrastive Pretraining for Fusion Plasma Diagnostics}

\author{Qiang Chen, Xiao Wang*, Qingquan Yang*, Hao Si, Zikang Yan, Guosheng Xu, Jin Tang
\thanks{$\bullet$ Qiang Chen, Xiao Wang, Hao Si, Zikang Yan, and Jin Tang are with the School of Computer Science and Technology, Anhui University, Hefei 230031, China (email: xiaowang@ahu.edu.cn).}       
\thanks{$\bullet$ Qingquan Yang and Guosheng Xu are with the Institute of Plasma Physics, Chinese Academy of Sciences, Hefei 230031, China (email: yangqq@ipp.ac.cn).} 
\thanks{* Corresponding Authors: Xiao Wang and Qingquan Yang.}  
}

\markboth{ IEEE Transactions on ***, 2026 } 
{Shell \MakeLowercase{\textit{et al.}}: Bare Demo of IEEEtran.cls for IEEE Journals}

\maketitle

\begin{abstract}
Visible-light videos and multivariate diagnostic signals provide
complementary observations of tokamak discharges. However, existing
fusion pretraining methods primarily learn representations by
reconstructing or predicting observations, leaving the correspondence
between these modalities underexplored. Representation learning is
further challenged by spatially localized plasma regions in video
frames and substantial differences in diagnostic sampling rates.
We propose FusionCLIP, a multimodal contrastive pretraining framework.
A selective video encoder retains informative visual tokens, while
a query-based time-series encoder integrates asynchronous diagnostic
signals using their observation timestamps. 
Contrastive learning aligns representations of the two modalities.
We construct the EAST Multimodal Diagnostics (EAST-MMD) dataset,
comprising 1307 EAST discharges, and evaluate the model through
independent fine-tuning on three tasks: disruption prediction,
edge-localized mode (ELM) classification, and $H_{98}$ regression.
FusionCLIP achieves a Joint score of 89.07\%, with absolute gains
of 2.41\% and 2.90\% over the best competing baseline, and the in-domain contrastive pretraining baseline CLIP-Base, respectively. Ablations and attention visualizations further support the contributions of its key components. 
The source code will be released on \url{https://github.com/Event-AHU/OpenFusion}
\end{abstract}


\section{Introduction}
\begin{figure*}[t]
    \centering
    \includegraphics[width=0.9\textwidth]{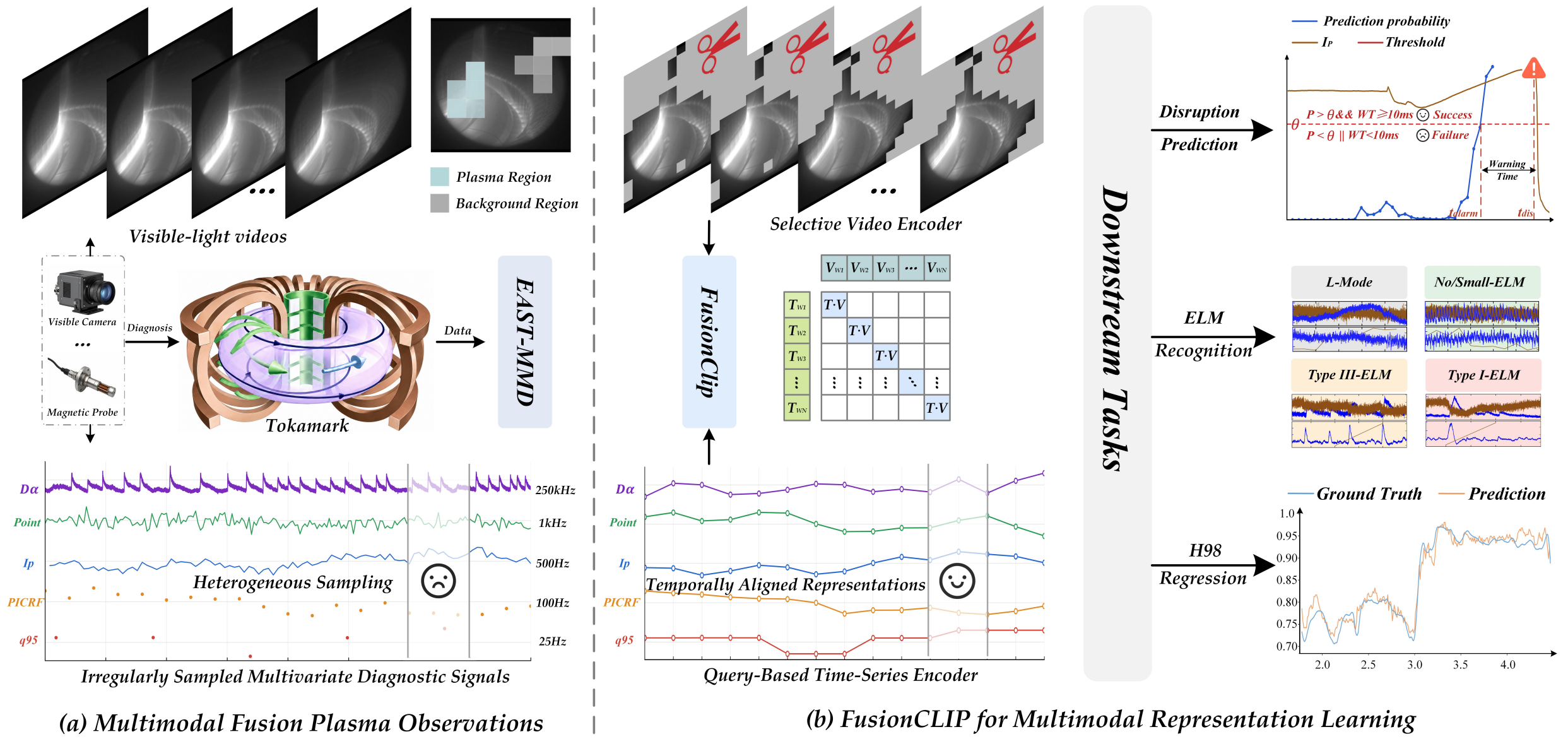}
    \caption{Overview of EAST-MMD and FusionCLIP. (a) Paired plasma videos with redundant backgrounds and diagnostic signals with heterogeneous sampling rates. (b) 
    Selective Video Encoder(SVE) and Query-Based Time-Series Encoder(QTE) encode the modalities for contrastive pretraining, followed by independent fine-tuning for disruption prediction, ELM classification, and $H_{98}$ regression.}   
    \label{fig:background}
\end{figure*}

Nuclear fusion has the potential to provide low-carbon energy, while stable tokamak operation requires continuous monitoring of high-temperature plasmas. During a tokamak discharge, different diagnostic systems observe the same plasma process from complementary perspectives. 
As shown in Figure~\ref{fig:background}(a), visible-light videos capture plasma spatial evolution, whereas multivariate diagnostic signals record the temporal evolution of different physical quantities. Learning transferable plasma representations from these observations is an important problem for intelligent fusion diagnostics.
Pretraining offers a way to learn transferable representations from large-scale data. 
In fusion research, FusionMAE~\cite{yang2026fusionmae} and TokaMind~\cite{boschi2026tokamind} have demonstrated the value of pretrained representations for downstream tasks through the reconstruction or prediction of observed data. 
Videos and diagnostic signals from the same time window form natural pairs, motivating cross-modal contrastive pretraining~\cite{radford2021clip} to learn representations transferable across fusion diagnostic tasks.
However, the distinct characteristics of the two modalities pose challenges to learning such representations.

As shown in Figure~\ref{fig:background}(a), plasma-related regions occupy only part of each frame amid extensive background and repeated observations.
Studies of visual representation learning suggest that how local information contributes to a representation matters. 
Registers~\cite{darcet2024registers} showed that background tokens may be repurposed for internal computation, while LaSt-ViT~\cite{shi2026lastvit} demonstrated that selective aggregation affects visual representations. 
Existing methods for vision-based fusion diagnostics~\cite{chen2026vidp,wang2026deltainvformer} primarily extract spatiotemporal features for specific diagnostic tasks. 
The role of local-region selection in learning transferable video representations remains less explored. 
Therefore, we explicitly select local information during video encoding, allowing the details of selected regions to be retained more fully.
For diagnostic time series, a central challenge is the heterogeneity of sampling frequencies. 
The signals in Figure~\ref{fig:background}(a) range from 25~Hz to 250~kHz. 
Existing fusion models often resample signals onto a fixed temporal grid~\cite{yang2023hl2a}.
A fixed grid simplifies input construction but may compress high-frequency variations and over-interpolate low-frequency signals, obscuring their original temporal structure.

To address these challenges, we propose FusionCLIP, a multimodal contrastive pretraining framework. 
As shown in Figure~\ref{fig:background}(b), the visual branch uses a Selective Video Encoder(SVE) to select tokens, preserving local details while reducing interference from background and repeated observations. Its temporal branch employs a Query-Based Time-Series Encoder(QTE), which uses temporal queries and time-biased attention to transform heterogeneously sampled signals into aligned representations.
The two branches are jointly pretrained with a bidirectional contrastive objective. 

We construct the EAST Multimodal Diagnostics (EAST-MMD) dataset and evaluate FusionCLIP on three independent tasks: disruption prediction, edge-localized mode (ELM) classification, and H98 regression. 
Compared with approaches using existing visual or time-series backbones, FusionCLIP achieves better results on multiple diagnostic metrics.\\
Our main contributions are as follows:
\begin{itemize}
    \item We introduce FusionCLIP, a multimodal contrastive pretraining framework for paired plasma videos and diagnostic signals.
    \item We introduce SVE and QTE to construct modality-specific representations that address the uneven distribution of local information in videos and the heterogeneous sampling of diagnostic signals, respectively.
    \item We construct EAST-MMD and evaluate the proposed pretraining framework and its key components on three independent diagnostic tasks.
\end{itemize}

\section{Related Work}
\subsection{Pretraining Paradigms}
Pretraining learns transferable representations from large-scale
unlabeled or weakly supervised data. By training objective, multimodal
pretraining includes masked modeling and reconstruction~\cite{geng2022multimodal,wang2023image,chen2024eve},
autoregressive modeling~\cite{wang2024emu3},
cross-modal contrastive learning~\cite{radford2021clip,jia2021scaling},
and multi-objective learning~\cite{li2021align,bao2022vlmo}.

In fusion, FusionMAE learns plasma-state representations by reconstructing
missing signals from multichannel HL-3 diagnostics~\cite{yang2026fusionmae}.
TokaMind learns transferable representations by reconstructing and
predicting heterogeneous MAST observations, including time series,
profiles, and videos~\cite{boschi2026tokamind}.
Bouchiat et al.~\cite{bouchiat2025foundation} explore masked autoencoding to integrate
DIII-D diagnostic and control data.

These fusion approaches primarily rely on reconstruction or prediction.
Cross-modal contrastive learning~\cite{radford2021clip,jia2021scaling}
exploits correspondence between paired observations but remains
underexplored for plasma videos and diagnostic signals.
FusionCLIP aligns videos and diagnostic signals within the same time window to learn transferable representations and evaluate contrastive pretraining for fusion diagnostics.

\subsection{Visual and Temporal Representation Learning}
Visual representation learning increasingly emphasizes local features
and their aggregation. DINO~\cite{caron2021dino},
Registers~\cite{darcet2024registers}, and LaSt-ViT~\cite{shi2026lastvit}
examine spatial semantics, background tokens, and selective aggregation,
respectively, highlighting the importance of local information organization.
Temporal models also account for sampling structure.
Moirai~\cite{woo2024moirai} and MOMENT~\cite{goswami2024moment}
model regular sequences on fixed time grids, while
GRU-D~\cite{che2018grud} and Raindrop~\cite{zhang2022raindrop}
incorporate temporal information to handle missing or irregular observations.

In fusion diagnostics, visible-light videos capture plasma evolution,
but salient plasma regions occupy only part of each frame.
Existing video models primarily target specific diagnostic
tasks~\cite{chen2026vidp,wang2026deltainvformer},
leaving local region selection for transferable representations
underexplored. Diagnostic signals also have heterogeneous sampling rates.
Existing approaches either organize signals on fixed time
grids~\cite{yang2023hl2a,croonen2023investigation}
or preserve sampling-rate differences across
diagnostics~\cite{zheng2023disruption,chen2026fusionmmt}.

Building on these studies, FusionCLIP uses spatiotemporal structure-guided
token selection to preserve plasma-related visual details and time-aware
queries to encode asynchronous, multi-rate diagnostic signals.
The resulting representations support contrastive pretraining
of paired videos and diagnostic signals.

\begin{figure*}[t]
    \centering
    \includegraphics[width=0.9\textwidth]{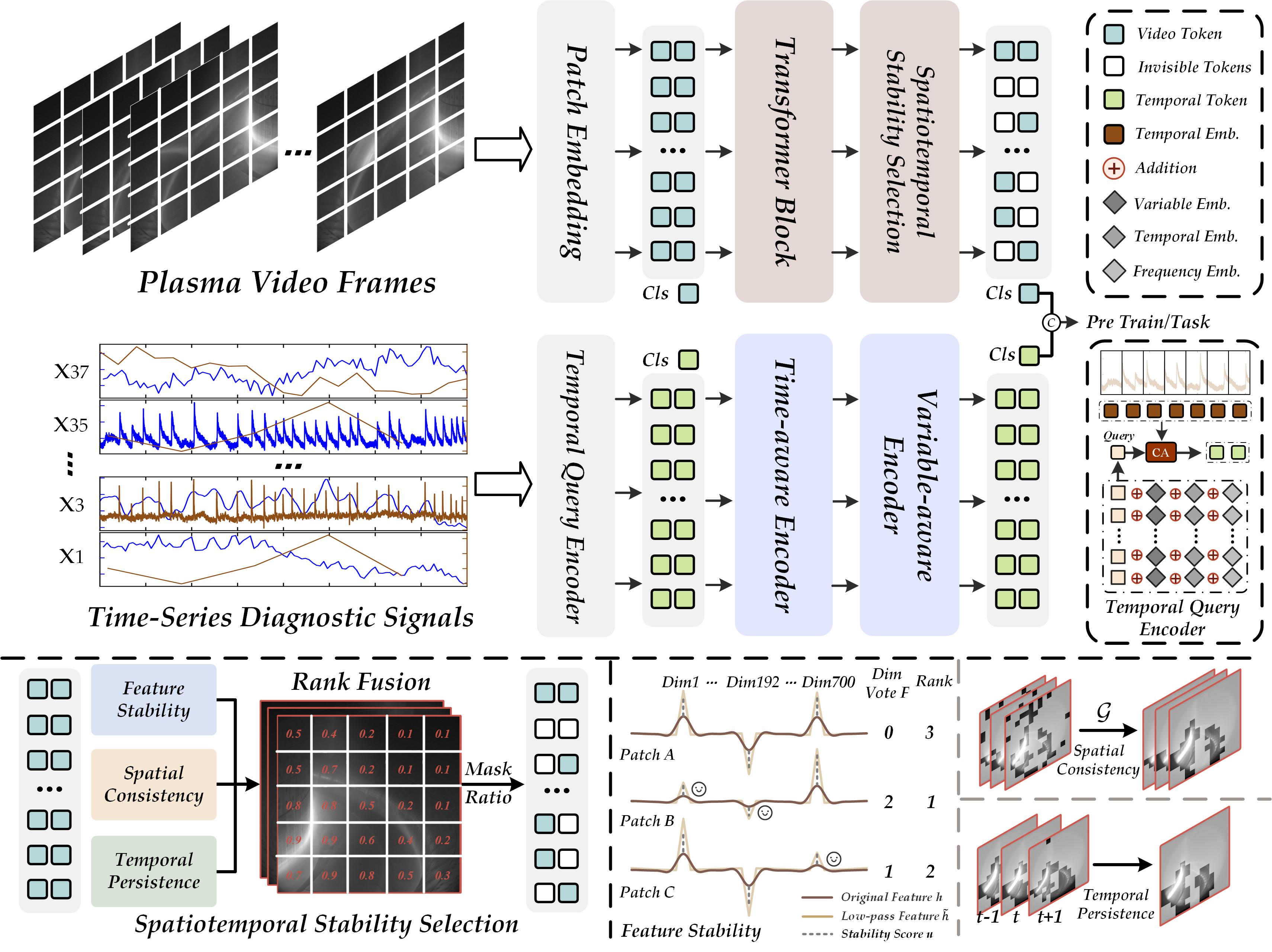}
   \caption{Overview of FusionCLIP. Selective video encoder selects video tokens by spatiotemporal stability. Query-based time-series encoder uses TQE to align irregularly sampled diagnostic signals onto a shared temporal grid and model temporal and inter-variable dependencies.}
    \label{fig:FusionCLIP}
\end{figure*}

\section{EAST Multimodal Diagnostics Dataset}
\subsection{Dataset Curation}
As shown in Fig.~\ref{fig:background}(a), we construct EAST Multimodal Diagnostics (EAST-MMD), comprising visible-light videos and multivariate diagnostic signals from 1307 EAST discharges.
667 discharges are used for multimodal pretraining and 640 for downstream evaluation. 
For pretraining, we use 50ms windows with a 2ms sliding stride, yielding 2725021 paired video-signal windows. 
The visible-light camera operates at 1 kHz, uniformly sampling 12 frames at a resolution of $224 \times 224$ from each window.
The diagnostic input contains 38 variables, each retaining its original observation times and sampling rate without resampling onto a common grid.
Downstream evaluation uses the task annotations and data splits of EAST-VTD640~\cite{chen2026fusionmmt}. This dataset contains 481 non-disruptive and 159 disruptive discharges, yielding 280094 annotated windows. The training, validation, and test sets are split by discharge and contain 447, 96, and 97 discharges, respectively.

\subsection{Downstream Tasks}
As illustrated in Fig.~\ref{fig:background}(b), we evaluate the pretrained representations on disruption prediction, ELM classification, and $H_{98}$ regression.
\paragraph{Disruption Prediction}
The model predicts disruption risk from the visible-light video and diagnostic signals in each window. At test time, windows from the same discharge are processed chronologically, and the first prediction to exceed a threshold selected on the validation set determines the alarm time. For a disruptive discharge, an alarm is counted as successful only if it precedes the disruption by at least 10ms. A threshold crossing on a non-disruptive discharge is counted as a false alarm.
\paragraph{ELM Classification}
The model identifies the plasma edge state associated with each window. The four classes are L-mode, No/Small ELM, Type-III ELM, and Type-I ELM.
\paragraph{$H_{98}$ Regression}
The model predicts a continuous $H_{98}$ value for each window to characterize the plasma energy confinement level.


\section{FusionCLIP}
\subsection{Overview}
For a given 50ms discharge window, let $\mathbf V_i$ denote the visible-light video and let the multivariate diagnostic time series be
$O_i=\{(o_{i,m,j},t_{i,m,j})\}_{m,j}$,
where $o_{i,m,j}$ is the $j$-th observation of diagnostic variable $m$ and $t_{i,m,j}$ is its actual timestamp. 
As shown in Figure~\ref{fig:FusionCLIP}, FusionCLIP extracts representations using a selective video encoder $E_v$ and a query-based time-series encoder $E_o$.
The video branch employs Spatiotemporal Stability Selection(SSS) to organize local visual information, while the temporal branch uses a Temporal Query Encoder(TQE) to read observations at their respective timestamps. Their outputs are denoted by
\begin{equation}
\mathbf h_{\mathrm{CLS},i}^{(v)}=E_v(\mathbf V_i),
\qquad
\mathbf h_{\mathrm{CLS},i}^{(o)}=E_o(\mathbf O_i).
\end{equation}
The two representations are used for multimodal contrastive pretraining and subsequently transferred to downstream diagnostic tasks.

\subsection{Selective Video Encoder}
Selective video encoder first converts consecutive frames into spatiotemporal patch tokens and uses Transformer interactions to form local features. 
Spatiotemporal Stability Selection combines feature stability, spatial consistency, and temporal persistence.
The selected tokens retain their individual representations, while the unselected tokens are aggregated into summary tokens. The CLS output is then used as the global video representation $\mathbf h_{\mathrm{CLS}}^{(v)}$.
Let $\mathbf h_{\tau,n}^{(v)}\in\mathbb R^D$ denote the visual feature at spatial position $n$ in temporal patch $\tau$. 
The normalized ranks of the three components are denoted by $R_{\tau,n}^{F}$, $R_{\tau,n}^{S}$, and $R_{\tau,n}^{T}$, respectively. 
As illustrated in Figure~\ref{fig:FusionCLIP}, rank fusion produces the final selection score:
\begin{equation}
P_{\tau,n}
= \lambda_F R_{\tau,n}^{F}
+ \lambda_S R_{\tau,n}^{S}
+ \lambda_T R_{\tau,n}^{T},
\end{equation}
where $\lambda_{F}$, $\lambda_{S}$, and $\lambda_{T}$ are weight hyperparameters.
The encoder retains high-scoring patches within each temporal block.

\begin{figure}[t]
    \centering
    \includegraphics[width=\columnwidth]{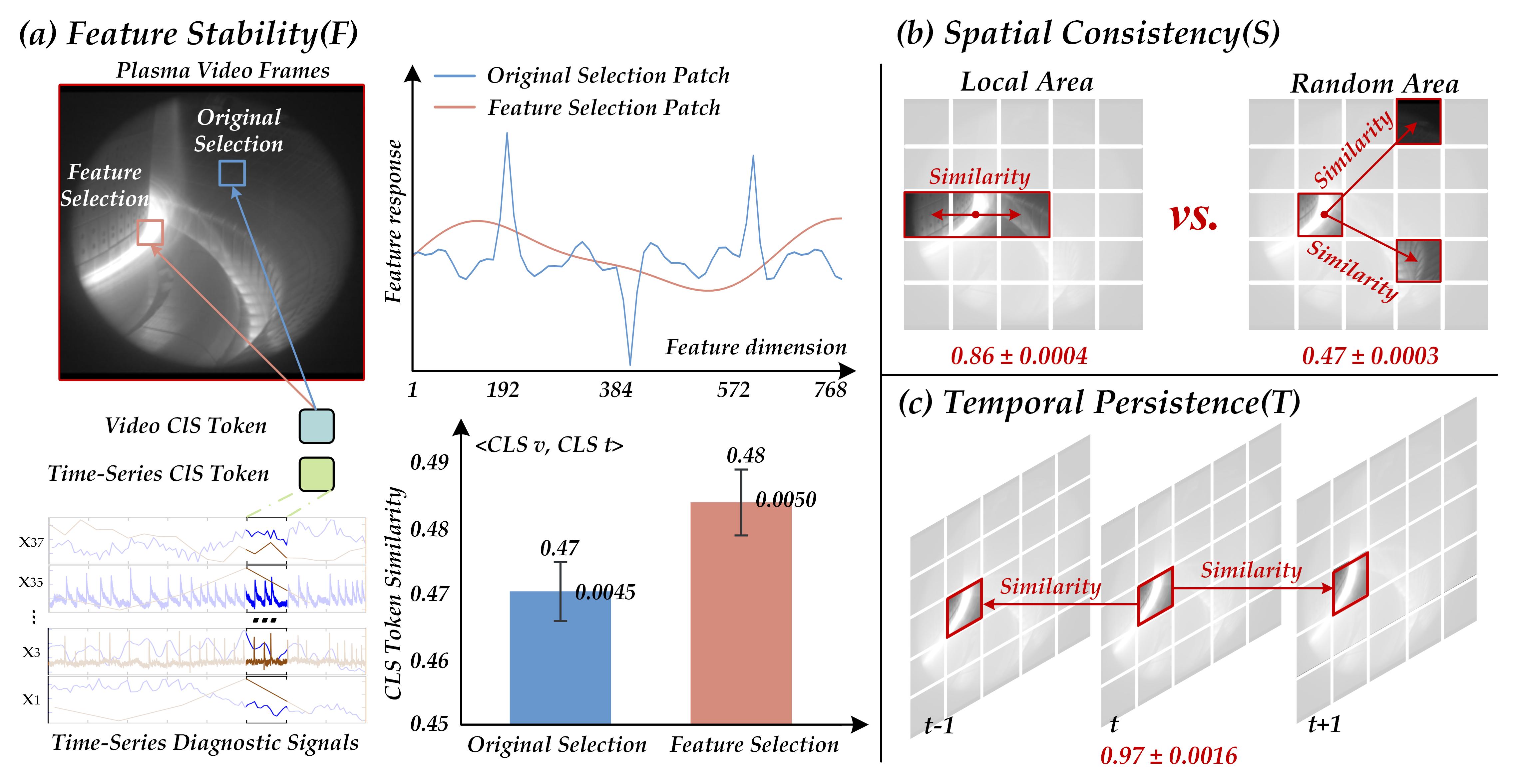}
    \caption{
    Feature Analysis for Spatiotemporal Stability Selection. Results are obtained using one sampled window per shot across five random seeds.
    (a) F favors patches with smoother responses, increasing video--time-series CLS similarity and suggesting stronger cross-modal correspondence.
    (b) Higher similarity between adjacent patches than random patches indicates spatial continuity.
    (c) A similarity of 0.97 between corresponding patches in adjacent tubelets indicates short-term temporal persistence.
    }
    \label{fig:Motivation_SSS}
\end{figure}

\paragraph{Feature Stability}
As shown in Fig.~\ref{fig:FusionCLIP} and Fig.~\ref{fig:Motivation_SSS}(a),
we hypothesize that after self-attention and contrastive pre-training, local visual features matching the corresponding window become more stable and thus exhibit smaller feature-wise response variations under high-frequency suppression. 
Based on this assumption, we construct a local retention priority by combining the observed response variation under high-frequency suppression with dimension-wise voting.
In Fig.~\ref{fig:Motivation_SSS}(a), we compare the similarity between the window video CLS token aggregated from the selected patches and the temporal CLS token with and without high-frequency suppression. The similarity increases from 0.47 to 0.48, further supporting our hypothesis.

As shown in Fig.~\ref{fig:FusionCLIP}, we compare each patch feature before and after filtering to guide local information selection.

Specifically, we apply a Fourier transform along the feature dimension of $\mathbf h_{\tau,n}^{(v)}$, suppress high-frequency components with $G$, and transform the result back:
\begin{equation}
\begin{aligned}
\widetilde{\mathbf h}_{\tau,n}^{(v)}
&=
\operatorname{irFFT}\!\left(
G\odot\operatorname{rFFT}(\mathbf h_{\tau,n}^{(v)})
\right),\\
G_k&=\exp\!\left[-\frac12\left(\frac{k}{\sqrt D}\right)^2\right].
\end{aligned}
\end{equation}
As the frequency index $k$ increases, $G_k$ decreases and progressively attenuates high-frequency components along the feature dimension. For feature dimension $d$, we compute the response relative to the filtering-induced change:
\begin{equation}
u_{\tau,n,d}
=
\frac{\left|h_{\tau,n,d}^{(v)}\right|}
{\max\!\left(
\left|\widetilde h_{\tau,n,d}^{(v)}
-h_{\tau,n,d}^{(v)}\right|,\epsilon
\right)}.
\end{equation}
where $\epsilon>0$ is a small constant that prevents division by zero.
Within each feature dimension, positions with smaller changes after high-frequency suppression score higher, suggesting that they may retain more plasma information aligned with the current temporal window.
At each temporal index $\tau$, we compare the stability responses of the $N$ spatial patches separately for each feature dimension $d$. Each dimension votes for the patch with the highest response, and $F_{\tau,n}$ denotes the fraction of votes received by patch $n$:
\begin{equation}
n_{\tau,d}^{*}=\underset{1\leq m\leq N}{\arg\max}\,u_{\tau,m,d},
\qquad
F_{\tau,n}=\frac{1}{D}\sum_{d=1}^{D}\mathbb I[n_{\tau,d}^{*}=n].
\end{equation}
Ranking $F_{\tau,n}$ across the spatial patches at temporal index $\tau$ yields the feature-stability rank $R_{\tau,n}^{F}$.

\paragraph{Spatial Consistency}
As shown in Figure~\ref{fig:FusionCLIP}, feature voting can highlight plasma-related patches, but high-scoring positions may remain spatially scattered. 
Fig.~\ref{fig:Motivation_SSS}(b) shows that neighboring patch similarity reaches 0.86, significantly higher than that between random regions. 
This indicates that visible-light plasma images exhibit local similarity. Therefore, we apply Gaussian spatial smoothing to the vote map:
$S=\mathcal G(\mathbf F)$, 
where $\mathcal G$ denotes Gaussian neighborhood smoothing. 
Ranking $\mathbf S$ yields the spatial-consistency rank $R^S$.

\paragraph{Temporal Persistence}
Fig.~\ref{fig:Motivation_SSS}(c) shows that patches at the same spatial location in adjacent temporal blocks have a similarity as high as 0.97, indicating persistent local plasma evolution. Therefore, we aggregate the responses at the same spatial location across adjacent temporal blocks using a 1:2:1 weighting:
\begin{equation}
C_{\tau,n}
=S_{\tau-1,n}+2S_{\tau,n}+S_{\tau+1,n}.
\end{equation}
Ranking $\mathbf C$ yields the temporal-consistency rank $R^T$. This component favors positions that receive support across neighboring time steps.

Finally, SSS selects visual tokens according to the fused score $P_{\tau,n}$. Unselected tokens within each temporal patch are averaged into one summary token. 
In subsequent encoding, the CLS token reads from the selected tokens and summary tokens, retaining local detail in selected regions while incorporating context from the remaining regions.

\begin{table*}[t]
    \centering
    \caption{
        Comparison on three downstream diagnostic tasks.
        TPR, FPR, F1, AUC, $R^2$, and Joint are reported in \%,
        AWT in ms, and MAE in its original scale.
        Best and second-best results are shown in \textbf{bold}
        and \underline{underlined}, respectively.
    }
    \label{tab:backbone_comparison_compact}

    \begingroup
    \fontsize{7.3}{8.5}\selectfont
    \setlength{\tabcolsep}{1.0pt}
    \setlength{\extrarowheight}{0.3pt}
    \renewcommand{\arraystretch}{1.15}

    \begin{tabular*}{\textwidth}{
        @{\extracolsep{\fill}}
        cccc
        *{7}{>{\centering\arraybackslash}p{23pt}}
        >{\centering\arraybackslash}p{31pt}
        >{\centering\arraybackslash}p{23pt}
        >{\centering\arraybackslash}p{24pt}
        @{}
    }
        \toprule

        \multirow{2}{*}{\textbf{Pattern}}
        & \multirow{2}{*}{\textbf{Visual}}
        & \multirow{2}{*}{\textbf{Time-series}}
        & \multirow{2}{*}{
            \makecell[c]{\textbf{Pretraining}\\\textbf{Dataset}}
        }
        & \multicolumn{5}{c}{\textbf{Disruption Prediction}}
        & \multicolumn{2}{c}{
            \makebox[0pt]{\textbf{ELM Classification}}
        }
        & \multicolumn{2}{c}{\textbf{H98 Regression}}
        & \multirow{2}{*}{\textbf{Joint}$\uparrow$} \\

        \cmidrule(lr){5-9}
        \cmidrule(lr){10-11}
        \cmidrule(lr){12-13}

        & & & &
        TPR$\uparrow$ & FPR$\downarrow$ & F1$\uparrow$
        & AUC$\uparrow$ & AWT$\uparrow$
        & F1$\uparrow$ & AUC$\uparrow$
        & MAE$\downarrow$ & $R^2\uparrow$ & \\

        \midrule

        Scratch
        & Transformer~\cite{vaswani2017attention}
        & Transformer
        & --
        & 70.83 & 4.11 & 77.27 & 91.32 & 103.12
        & 77.24 & \underline{97.74}
        & 0.026590 & 92.58 & 82.36 \\

        \midrule

        \multirow{4}{*}[-10.8pt]{\makecell[c]{Visual / QTE}}
        & MAE~\cite{he2022mae}
        & QTE
        & \makecell[c]{ImageNet-1K\\EAST-MMD}
        & 87.50 & 4.11 & 87.50 & 97.03 & 53.10
        & 78.13 & 97.52
        & \underline{0.025910} & \underline{93.14} & 86.26 \\

        & DINO~\cite{caron2021dino}
        & QTE
        & \makecell[c]{ImageNet-1K\\EAST-MMD}
        & \underline{91.67} & 12.33 & 80.00 & 95.89 & \textbf{348.27}
        & \underline{78.14} & 97.56
        & 0.028170 & 92.53 & 83.56 \\

        & CLIP~\cite{radford2021clip}
        & QTE
        & \makecell[c]{WIT-400M\\EAST-MMD}
        & 83.33 & 4.11 & 85.11 & 97.17 & 118.60
        & 78.00 & 97.15
        & 0.026190 & 93.13 & 85.41 \\

        & MetaCLIP~\cite{xu2024metaclip}
        & QTE
        & \makecell[c]{MetaCLIP-400M\\EAST-MMD}
        & 83.33 & 4.11 & 85.11 & 97.00 & 69.20
        & 77.65 & 97.56
        & 0.028020 & 92.32 & 85.03 \\

        \midrule

        \multirow{3}{*}[-7.3pt]{
            \makecell[c]{SVE / Regular\\Time Series}
        }
        & SVE
        & TSPulse~\cite{ekambaram2026tspulse}
        & \makecell[c]{EAST-MMD\\Monash/LibCity}
        & 83.33 & \underline{1.37} & 88.89 & 97.72 & 87.20
        & 76.26 & 96.97
        & 0.053366 & 77.34 & 80.83 \\

        & SVE
        & Moirai~\cite{woo2024moirai}
        & \makecell[c]{EAST-MMD\\LOTSA}
        & 87.50 & \textbf{0.00} & \underline{93.33} & 98.40 & 66.81
        & 76.94 & 97.28
        & 0.055868 & 75.97 & 82.08 \\

        & SVE
        & MOMENT~\cite{goswami2024moment}
        & \makecell[c]{EAST-MMD\\TS-Pile}
        & \underline{91.67} & 2.74 & 91.67 & 97.63 & 50.64
        & 75.94 & 97.09
        & 0.051621 & 78.28 & 81.96 \\

        \midrule

        \multirow{4}{*}[-10.8pt]{
            \makecell[c]{SVE / Irregular\\Time Series}
        }
        & SVE
        & ISTS-PLM~\cite{zhang2025istsplm}
        & \makecell[c]{EAST-MMD\\--}
        & \underline{91.67} & 4.11 & 89.80 & 97.69 & 91.95
        & 74.23 & 97.44
        & 0.030808 & 91.43 & 85.15 \\

        & SVE
        & Hi-Patch~\cite{luo2025hipatch}
        & \makecell[c]{EAST-MMD\\--}
        & \underline{91.67} & 2.74 & 91.67 & 98.52 & 140.45
        & 75.97 & 97.62
        & 0.026736 & 92.35 & \underline{86.66} \\

        & SVE
        & GRU-D~\cite{che2018grud}
        & \makecell[c]{EAST-MMD\\--}
        & \textbf{95.83} & 5.48 & 90.20 & 98.17 & 151.61
        & 72.14 & 96.69
        & 0.033058 & 88.23 & 83.52 \\

        & SVE
        & Raindrop~\cite{zhang2022raindrop}
        & \makecell[c]{EAST-MMD\\--}
        & 79.17 & 4.11 & 82.61 & 95.23 & 111.63
        & 75.54 & 96.51
        & 0.032887 & 90.21 & 82.79 \\

        \midrule

        CLIP-Base
        & MAE
        & Moirai
        & EAST-MMD
        & \underline{91.67} & 4.11 & 89.80 & \textbf{98.69} & 103.73
        & 78.11 & \textbf{97.91}
        & 0.029495 & 90.61 & 86.17 \\

        FusionCLIP
        & SVE
        & QTE
        & EAST-MMD
        & \textbf{95.83} & 2.74 & \textbf{93.88}
        & \underline{98.63} & \underline{222.17}
        & \textbf{79.82} & 97.61
        & \textbf{0.025362} & \textbf{93.50} & \textbf{89.07} \\

        \bottomrule
    \end{tabular*}
    \endgroup
\end{table*}

\subsection{Query-Based Time-Series Encoder}
Fusion diagnostic variables are sampled at substantially different frequencies.
Resampling them onto a fixed grid may suppress high-frequency variations or
introduce extensive interpolation for low-frequency signals. To preserve observation time information, we employ a temporal query encoder (TQE) to construct time-series representations and further model temporal features along the temporal and variable dimensions.

TQE encodes low-frequency variables observation by observation and
high-frequency variables patch by patch. Patch-based encoding first reduces
the number of events produced by densely sampled signals and the cost of
subsequent processing. 
Let $\mathbf p_{i,m,k}$ denote the $k$-th observation patch of
high-frequency variable $m$ in window $i$, with center time
$\tilde t_{i,m,k}$. For low-frequency observations,
$\tilde t_{i,m,j}$ is the original timestamp.
The two types of event features are
\begin{equation}
\begin{aligned}
\mathbf e_{i,m,j}^{(L)}
&=f_L(o_{i,m,j})+\phi(\tilde t_{i,m,j})
+\mathbf v_m+\mathbf c_L,\\
\mathbf e_{i,m,k}^{(H)}
&=f_H(\mathbf p_{i,m,k})+\phi(\tilde t_{i,m,k})
+\mathbf v_m+\mathbf c_H.
\end{aligned}
\end{equation}
Functions $f_L$ and $f_H$ embed individual observations and patches,respectively. $\phi$ encodes time, $\mathbf v_m$ is the variable embedding, and $\mathbf c_L$ and $\mathbf c_H$ are the respective low- and high-frequency type embeddings.
Below, $\mathbf e_{i,m,r}$ and $\tilde t_{i,m,r}$ denote either
type of encoded event and its associated time.

Within each 50-ms window, TQE uses 25 learnable time queries shared across
variables. Adding the query time $s_\ell$ and a variable embedding gives
\begin{equation}
\mathbf q_{i,m,\ell}
=\mathbf q_\ell+\phi(s_\ell)+\mathbf v_m,
\qquad \ell=1,\ldots,25.
\end{equation}
Each query attends to the events of its corresponding variable. Although
both queries and events contain time embeddings, attention may still assign
high weights to temporally distant observations with similar features.
We therefore add a time-distance bias to encourage attention to nearby
signal changes. For attention head $h$, the logit is
\begin{equation}
\begin{aligned}
a_{i,m,\ell r}^{(h)}
&=
\frac{
(\mathbf W_Q^{(h)}\mathbf q_{i,m,\ell})^\top
(\mathbf W_K^{(h)}\mathbf e_{i,m,r})
}{\sqrt{d_h}}\\
&\quad{}-\lambda_h\left|s_\ell-\tilde t_{i,m,r}\right|,\\
\lambda_h&=\operatorname{softplus}(\gamma_h),
\end{aligned}
\end{equation}
where $d_h$ is the attention-head dimension and $\lambda_h$ controls
the strength of the time-distance bias. Attention weights are normalized
over valid events, allowing variables with different sampling rates to
form representations at shared query times. A temporal backbone then
models dependencies within each variable, followed by cross-variable
attention. Finally, the resulting features are aggregated into the
global representation $\mathbf h_{\mathrm{CLS},i}^{(o)}$.

\providecommand{\yes}{\ding{51}}
\providecommand{\no}{\ding{55}}

\begin{table*}[t]
    \centering
    \caption{
        Ablation study of SSS in the SVE, the TQE in the QTE, and multimodal pretraining.
        TPR, FPR, F1, AUC, $R^2$, and Joint are reported in \%,
        AWT in ms, and MAE in its original scale.
        Best and second-best results are shown in \textbf{bold}
        and \underline{underlined}, respectively.
    }
    \label{tab:component_ablation}

    \begingroup
    \fontsize{7.3}{8.5}\selectfont
    \setlength{\tabcolsep}{3.5pt}
    \renewcommand{\arraystretch}{1.12}

    \begin{tabular*}{\textwidth}{
        @{\extracolsep{\fill}}
        ccc*{10}{c}
        @{}
    }
        \toprule

        \multirow{2}{*}{\textbf{Pre-training}}
        & \multirow{2}{*}{\textbf{SSS}}
        & \multirow{2}{*}{\textbf{TQE}}
        & \multicolumn{5}{c}{\textbf{Disruption Prediction}}
        & \multicolumn{2}{c}{\textbf{ELM Classification}}
        & \multicolumn{2}{c}{\textbf{H98 Regression}}
        & \multirow{2}{*}{\textbf{Joint}$\uparrow$} \\

        \cmidrule(lr){4-8}
        \cmidrule(lr){9-10}
        \cmidrule(lr){11-12}

        & & &
        \textbf{TPR}$\uparrow$
        & \textbf{FPR}$\downarrow$
        & \textbf{F1}$\uparrow$
        & \textbf{AUC}$\uparrow$
        & \textbf{AWT}$\uparrow$
        & \textbf{F1}$\uparrow$
        & \textbf{AUC}$\uparrow$
        & \textbf{MAE}$\downarrow$
        & $\mathbf{R}^{2}\uparrow$
        & \\

        \midrule

        \no & \no & \no
        & \underline{87.50} & \textbf{1.37} & 91.30
        & \textbf{99.20} & 77.48
        & 75.82 & 97.35
        & 0.031259 & 89.44 & 85.52 \\

        \no & \no & \yes
        & 83.33 & \underline{2.74} & 86.96
        & 98.06 & 73.85
        & 76.25 & \underline{97.81}
        & \underline{0.026874} & \underline{93.13} & 85.44 \\

        \no & \yes & \yes
        & \textbf{95.83} & 4.11 & \underline{92.00}
        & 97.43 & \underline{96.39}
        & \underline{77.21} & \textbf{97.86}
        & 0.030168 & 92.23 & \underline{87.15} \\

        \yes & \yes & \yes
        & \textbf{95.83} & \underline{2.74} & \textbf{93.88}
        & \underline{98.63} & \textbf{222.17}
        & \textbf{79.82} & 97.61
        & \textbf{0.025362} & \textbf{93.50} & \textbf{89.07} \\

        \bottomrule
    \end{tabular*}
    \endgroup
\end{table*}

\begin{table*}[t]
    \centering
    \caption{
        Effects of visual token selection and mask ratio on diagnostic performance.
        Mask ratios, TPR, FPR, F1, AUC, $R^2$, and Joint are reported in \%,
        AWT in ms, and MAE in its original scale.
        Best and second-best results are shown in \textbf{bold}
        and \underline{underlined}, respectively.
    }
    \label{tab:mask_ratio}

    \begingroup
    \fontsize{7.3}{8.5}\selectfont
    \setlength{\tabcolsep}{3.5pt}
    \renewcommand{\arraystretch}{1.12}

    \begin{tabular*}{\textwidth}{
        @{\extracolsep{\fill}}
        cc*{10}{c}
        @{}
    }
        \toprule

        \multirow{2}{*}{\textbf{Module}}
        & \multirow{2}{*}{\textbf{Mask Ratio}}
        & \multicolumn{5}{c}{\textbf{Disruption Prediction}}
        & \multicolumn{2}{c}{\textbf{ELM Classification}}
        & \multicolumn{2}{c}{\textbf{H98 Regression}}
        & \multirow{2}{*}{\textbf{Joint}$\uparrow$} \\

        \cmidrule(lr){3-7}
        \cmidrule(lr){8-9}
        \cmidrule(lr){10-11}

        & &
        \textbf{TPR}$\uparrow$
        & \textbf{FPR}$\downarrow$
        & \textbf{F1}$\uparrow$
        & \textbf{AUC}$\uparrow$
        & \textbf{AWT}$\uparrow$
        & \textbf{F1}$\uparrow$
        & \textbf{AUC}$\uparrow$
        & \textbf{MAE}$\downarrow$
        & $\mathbf{R}^{2}\uparrow$
        & \\

        \midrule

        Time Series & --
        & 83.33 & \underline{4.11} & 85.11
        & 96.40 & 47.1
        & 77.04 & 97.71
        & 0.02576 & 93.40 & 85.18 \\

        Video & SSS 50
        & \underline{91.67} & \textbf{2.74} & \underline{91.67}
        & \textbf{98.86} & 73.82
        & 68.04 & 95.78
        & 0.059 & 72.21 & 77.31 \\

        \midrule

        \multirow{5}{*}{Multimodal} & Full Token
        & \underline{91.67} & 5.48 & 88.00
        & 95.49 & 149.41
        & \underline{79.18} & \textbf{97.82}
        & 0.025787 & \underline{93.68} & \underline{86.95} \\

        & Random 50
        & \underline{91.67} & 5.48 & 88.00
        & 96.00 & 60.05
        & 77.67 & 97.46
        & 0.034071 & 91.09 & 85.59 \\

        & SSS 25
        & 79.17 & \textbf{2.74} & 84.44
        & 96.52 & \underline{156.11}
        & 79.08 & \underline{97.76}
        & \textbf{0.023024} & \textbf{94.11} & 85.88 \\

        & SSS 50
        & \textbf{95.83} & \textbf{2.74} & \textbf{93.88}
        & \underline{98.63} & \textbf{222.17}
        & \textbf{79.82} & 97.61
        & \underline{0.025362} & 93.50 & \textbf{89.07} \\

        & SSS 75
        & 87.50 & 5.48 & 85.71
        & 96.38 & 149.38
        & 77.61 & 97.29
        & 0.026311 & 93.22 & 85.51 \\

        \bottomrule
    \end{tabular*}
    \endgroup
\end{table*}

\subsection{Multimodal Contrastive Pretraining}
The video and diagnostic time series from the same window, $(\mathbf V_i,\mathbf O_i)$, form a positive cross-modal pair. We project the global representations of the two encoders into a shared space. For $a\in\{v,o\}$, let $\mathbf z_i^{(a)}=\operatorname{norm}(\mathbf W_a\mathbf h_{\mathrm{CLS},i}^{(a)})$ be the normalized projection. The cross-modal similarity is $d_{ij}=(\mathbf z_i^{(v)})^\top\mathbf z_j^{(o)}/\theta$, where the learnable temperature $\theta$ is initialized to $0.07$.

Let $\mathcal{I}$ index valid paired samples and
$\mathcal{A}_i \subseteq \mathcal{I}$ contain the same-window
positive index $i$ and negative indices from other windows.
The video-to-time-series and time-series-to-video losses are
defined as:
\begin{equation}
\begin{aligned}
\mathcal L_{v\to o}
&=-\frac{1}{|\mathcal I|}\sum_{i\in\mathcal I}
\log\frac{e^{d_{ii}}}{\sum_{j\in\mathcal A_i}e^{d_{ij}}},\\
\mathcal L_{o\to v}
&=-\frac{1}{|\mathcal I|}\sum_{i\in\mathcal I}
\log\frac{e^{d_{ii}}}{\sum_{j\in\mathcal A_i}e^{d_{ji}}}.
\end{aligned}
\end{equation}
The pretraining objective averages the two directions:
\begin{equation}
\mathcal L_{\mathrm{CLIP}}
=\frac{\mathcal L_{v\to o}+\mathcal L_{o\to v}}{2}.
\end{equation}
This objective trains the encoders to capture the correspondence between paired videos and diagnostic time series. After pretraining, the encoders are transferred to downstream diagnostic tasks and fine-tuned independently.




\section{Experiments}
\subsection{Evaluation Metrics and Experimental Setup}
\noindent\paragraph{Evaluation Metrics}
The three downstream tasks have different output forms, and we adopt task-specific evaluation metrics. For disruption prediction, we use discharge-level evaluation and report TPR, FPR, F1, AUC, and average warning time (AWT). For ELM classification, we report Macro-F1 and Macro-AUC. For $H_{98}$ regression, we use MAE and $R^2$. The definitions of AWT and the overall Joint metric are given as:
\begin{equation}
\begin{aligned}
AWT&=\frac{1}{N_s}\sum_{i=1}^{N_s}(t_{dis}^{i}-t_{alarm}^{i}),\\
Joint&=\frac{F1_{dis}+F1_{elm}+R^2_{H98}}{3}.
\end{aligned}
\end{equation}
$N_s$ counts disruptive discharges with successful warnings. Higher TPR, F1, AUC, AWT, and $R^2$, and lower FPR and MAE, indicate better performance.

\noindent\paragraph{Experimental Setup}
We pretrain FusionCLIP on EAST-MMD for 10 epochs using AdamW with a batch size of 256, a learning rate of $3\times10^{-5}$, and a weight decay of 0.05. 
Each sample comprises a 12-frame video and diagnostic signals from 38 variables, with a visual token masking ratio of $50\%$.
Contrastive learning uses a 256-dimensional embedding and a temperature of 0.07. We fine-tune the pretrained model separately for each downstream task for 10 epochs, using a batch size of 128 and a learning rate of $5\times10^{-5}$, while keeping the weight decay and mask ratio unchanged. Pretraining uses two NVIDIA A800 GPUs, and fine-tuning uses one.
The weight hyperparameters are $(\lambda_F, \lambda_S, \lambda_T) = (0.20, 0.35, 0.45)$.
The visual and temporal encoders are initialized with pretrained MAE and Moirai weights, respectively.

\begin{figure*}[t]
    \centering
    \includegraphics[width=\textwidth]{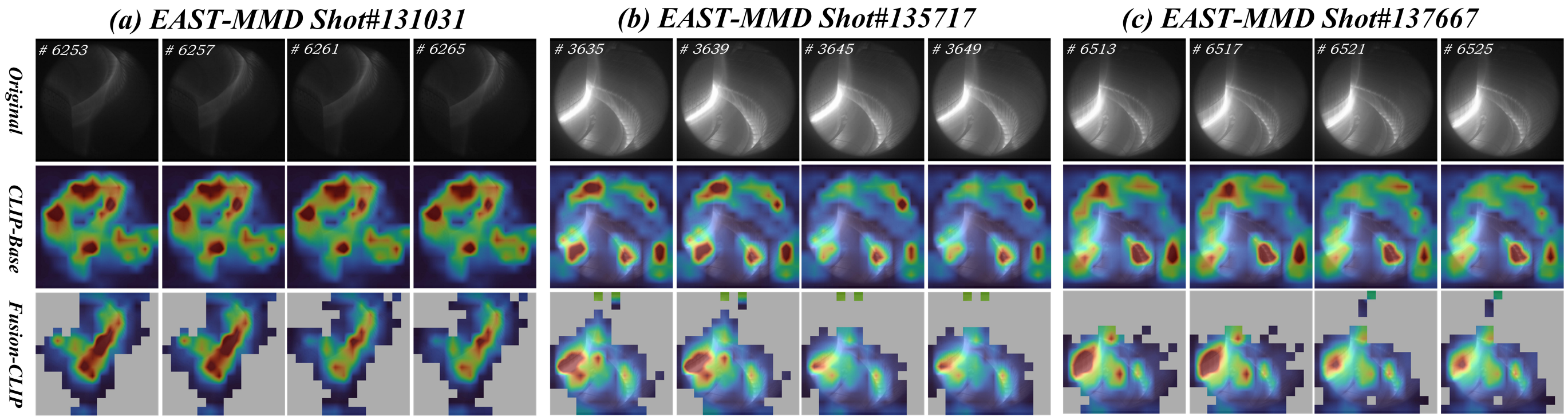}
    \caption{CLS-to-patch attention distributions of pretrained CLIP-Base and FusionCLIP models on EAST-MMD. The first row shows the original video frames. The second and third rows show the pretrained attention-weight maps of CLIP-Base and FusionCLIP, respectively.}
    \label{fig:cls_attion}
\end{figure*}

\begin{figure*}[t]
    \centering
    \includegraphics[width=\textwidth]{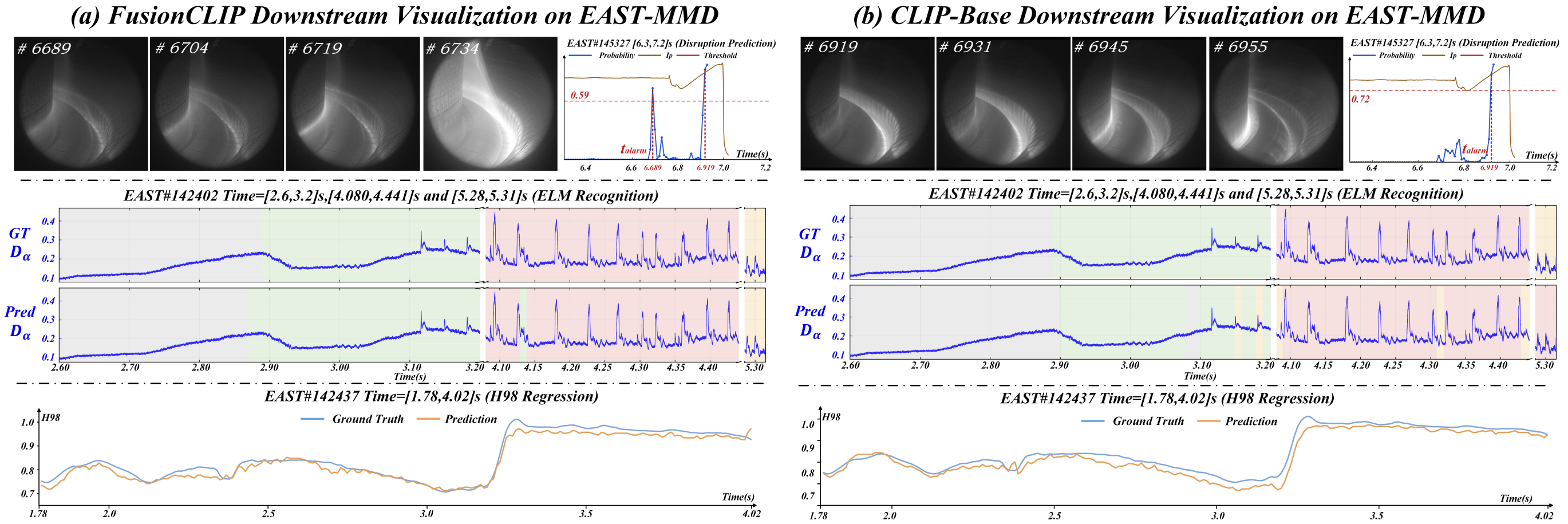}
    \caption{Downstream prediction examples for disruption prediction, ELM classification, and H98 regression on the EAST-MMD dataset. (a) Predictions from FusionCLIP. (b) Predictions from CLIP-Base on the corresponding examples. The x-axis of the disruption curves represents the starting time of the input window. 6.689 s indicates an input window interval of $[6.689, 6.739]$ s.}
    \label{fig:dw_result}
\end{figure*}

\subsection{Comparison with Pretrained Models}
Table 1 compares four baselines: training from scratch, general visual pre-training, general regular time-series pre-training, and general irregular time-series backbones. 
For a fair comparison, we combine the pre-trained visual/time-series backbones with general time-series/visual backbones to jointly optimize downstream tasks.
CLIP-Base denotes the pre-trained version without SSS for visual token selection or TQE for encoding.
All models were trained on the original EAST-MMD dataset. 

As shown in Table~\ref{tab:backbone_comparison_compact}, FusionCLIP achieves the highest Joint score of 89.07\%, with absolute gains of 6.71\%, 2.81\%, 6.99\%, and 2.41\% over Scratch and the best visual (MAE/QTE), regular time-series (SVE/Moirai), and irregular time-series (SVE/Hi-Patch) configurations, respectively. It also achieves an absolute gain of 2.90\% over the in-domain contrastive pretraining baseline CLIP-Base, demonstrating its overall advantage across the three downstream tasks.
Table~\ref{tab:mask_ratio} shows that time series perform better on ELM classification and H98 regression, whereas video performs better on disruption prediction. As shown in Table~\ref{tab:backbone_comparison_compact}, Visual/QTE configurations using our pretrained QTE generally outperform alternative temporal backbones on ELM classification and H98 regression, while most configurations using our pretrained SVE achieve higher disruption F1 than Visual/QTE. These results support the representation capabilities of both encoders in their respective modalities, with their combination in FusionCLIP yielding the best overall performance.

\subsection{Ablations and Analysis}
\noindent\paragraph{Pretraining and Component Analysis}
We conduct ablation studies on SSS, the core module of SVE, and TQE, the core module of QTE.
Table~\ref{tab:component_ablation} shows that adding TQE improves ELM classification and H98 regression but reduces disruption F1. Together with the unimodal results in Table~\ref{tab:mask_ratio}, this suggests that TQE may better capture temporal representations relevant to ELM and H98. 
Adding SSS raises disruption F1 and ELM F1 to 92.00\% and 77.21\%, respectively. 
These gains suggest that SSS may preserve features compatible with the time-series modality. 
Contrastive pretraining further improves F1 on both classification tasks and reduces H98 MAE, raising the Joint score to 89.07\% and supporting the benefit of cross-modal alignment for downstream diagnostics.

\noindent\paragraph{Visual Token Analysis}
Table~\ref{tab:mask_ratio} shows that video performs better on disruption prediction, whereas time series performs better on ELM classification and H98 regression. 
All multimodal configurations outperform both unimodal baselines in Joint score, indicating complementary information across modalities. 
SSS with a 50\% mask ratio achieves the highest Joint score of 89.07\%, exceeding full tokens and random selection at the same mask ratio by 2.12\% and 3.48\%, respectively. 
Although a 25\% mask ratio performs best on H98 regression, the 50\% setting achieves higher disruption F1 and ELM F1 and is therefore adopted as the default.

\subsection{Visualization Analysis}
\noindent\paragraph{Pretraining Attention Visualization}
Figure~\ref{fig:cls_attion} compares the CLS-to-patch attention maps of pretrained CLIP-Base and FusionCLIP. CLIP-Base assigns substantial attention to background regions, whereas FusionCLIP masks many background patches and concentrates its attention more strongly on visible plasma structures. This qualitative difference suggests that SSS helps preserve informative local regions when forming the pretrained video representation.

\noindent\paragraph{Downstream Prediction Examples}
Figure~\ref{fig:dw_result} compares FusionCLIP and CLIP-Base on three downstream tasks. 
For discharge \#145327, a visible abnormality appears near 6.734s.
FusionCLIP flags the window starting at 6.689s, while CLIP-Base
remains below the warning threshold for the same window.
FusionCLIP responds again to a subsequent abnormality near 6.919s. The ELM and H98 panels show additional prediction examples from both models.

\section{Conclusion}
We presented FusionCLIP to learn transferable representations
from paired videos and diagnostic signals. To address visual
redundancy and heterogeneous sampling, SVE uses SSS to retain
informative tokens, while QTE integrates signals using observation
timestamps. Contrastive pretraining aligns both modalities.
On EAST-MMD, FusionCLIP achieves the highest Joint score of
89.07\% among the evaluated configurations across three independently
fine-tuned tasks. Ablations support the contributions of pretraining
and the proposed components. Unimodal comparisons favor video
for disruption prediction and signals for ELM classification
and $H_{98}$ regression. Attention visualizations show that
FusionCLIP focuses more strongly on plasma structures than
CLIP-Base. Future work will extend the current EAST-only
evaluation to cross-tokamak transfer.


\small{ 
\bibliographystyle{IEEEtran}
\bibliography{main}

}


\end{document}